\documentclass[a4paper,fleqn]{cas-dc}
 \pdfoutput=1

\usepackage[authoryear,longnamesfirst]{natbib}
\usepackage{booktabs}
\usepackage{bbding}
\usepackage{graphicx}
\usepackage{subfigure}

\def\tsc#1{\csdef{#1}{\textsc{\lowercase{#1}}\xspace}}
\tsc{WGM}
\tsc{QE}

\begin{document}
\let\WriteBookmarks\relax
\def\floatpagepagefraction{1}
\def\textpagefraction{.001}

\shorttitle{MMA-RNN for AF discrimination and localization}    

\shortauthors{Y Sun et~al.}  

\title [mode = title]{MMA-RNN: A Multi-level Multi-task Attention-based Recurrent Neural Network for Discrimination and Localization of Atrial Fibrillation}  



%

\author[1,2]{Yifan Sun}[
                        ]
\fnmark[1]
\author[1,2]{Jingyan Shen}
\fnmark[1]
\author[1,2]{Yunfan Jiang}
\author[3]{Zhaohui Huang}
\author[4]{Minsheng Hao}
\author[4,5]{Xuegong Zhang}[
 orcid=0000-0002-9684-5643
                        ]
    
\cormark[1]

\ead{zhangxg@tsinghua.edu.cn}

\ead[url]{https://www.au.tsinghua.edu.cn/info/1110/1569.htm}


\affiliation[1]{organization={Department of Management Science and Engineering, Tsinghua University},
            city={Beijing},
            postcode={100084}, 
            country={China}}
\affiliation[2]{organization={Department of Industrial Engineering and Operations Research, Columbia University},
            city={New York},
            postcode={10027}, 
            country={United States}}
\affiliation[3]{organization={Department of Electronic Engineering, Tsinghua University},
            city={Beijing},
            postcode={100084}, 
            country={China}}
\affiliation[4]{organization={MOE Key Laboratory of Bioinformatics and Bioinformatics Division, BNRIST, Department of Automation, Tsinghua University},
            city={Beijing},
            postcode={100084}, 
            country={China}}
\affiliation[5]{organization={School of Life Sciences, Center for Synthetic and Systems Biology, Tsinghua University},
            city={Beijing},
            postcode={100084}, 
            country={China}}






\cortext[1]{Corresponding author}

\fntext[1]{These authors contributed equally to this work.}


\begin{abstract}
The automatic detection of atrial fibrillation based on electrocardiograph (ECG) signals has received wide attention both clinically and practically. It is challenging to process ECG signals with cyclical pattern, varying length and unstable quality due to noise and distortion. Besides, there has been insufficient research on separating persistent atrial fibrillation from paroxysmal atrial fibrillation, and little discussion on locating the onsets and end points of AF episodes. It is even more arduous to perform well on these two distinct but interrelated tasks, while avoiding the mistakes inherent from stage-by-stage approaches.
This paper proposes the Multi-level Multi-task Attention-based Recurrent Neural Network for three-class discrimination on patients and localization of the exact timing of AF episodes. Our model captures three-level sequential features based on a hierarchical architecture utilizing Bidirectional Long and Short-Term Memory Network (Bi-LSTM) and attention layers, and accomplishes the two tasks simultaneously with a multi-head classifier. The model is designed as an end-to-end framework to enhance information interaction and reduce error accumulation. Finally, we conduct experiments on CPSC 2021 dataset and the result demonstrates the superior performance of our method, indicating the potential application of MMA-RNN to wearable mobile devices for routine AF monitoring and early diagnosis. 
\end{abstract}




\maketitle

\section{Introduction}

\begin{figure*}[!tpb]
\setlength{\abovecaptionskip}{0.cm} 
\centerline{\includegraphics[width=1\textwidth]{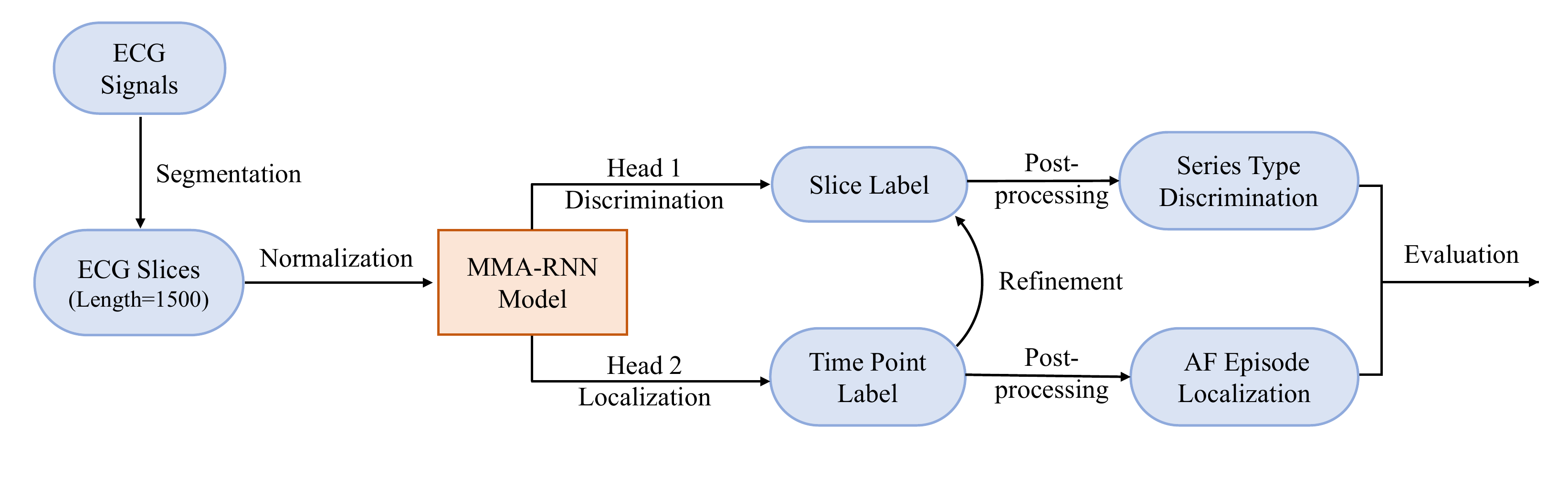}}
\caption{Pipeline of study. The main processing steps include segmentation, normalization, obtaining multi-head outputs (MMA-RNN model for both discrimination and localization task), post-processing, and final result evaluation.
}\label{pipeline}
\end{figure*}

Atrial fibrillation (AF) is one of the most common sustained arrhythmias with an irregular and often very rapid heart rhythm. Although AF itself usually isn't life-threatening, it can lead to severe complications such as cardiac failure and atrial thrombosis \citep{Larburu2011ComparativeSO}. Clinically, electrocardiogram (ECG), the noninvasive graphical visualization test that records the electrical activity of the heart, is a common tool to detect AF \citep{Larburu2011ComparativeSO}. However, the correct interpretation of ECG signals requires sufficient experts' professional knowledge and clinical experience. Therefore, many data-based methods identifying changes in cardiac electrical signals have been proposed \citep{Kalidas2019DetectionOA, Ghiasi2017}. Recently, there is a rapid rise of wearable mobile ECG devices, suggesting that more explorations on automatic AF detection algorithms to fulfill monitoring and diagnosis of the disease are indispensable \citep{isakadze2020useful, sion2021AI}.

Early AF detection works mainly focused on the binary discrimination problem, i.e., distinguishing between normal ($N$) and AF patients. Whereas, AF can be further divided into continuous (persistent atrial fibrillation, $AF_f$) and intermittent (paroxysmal atrial fibrillation, $AF_p$) types, which requires different patient management and specific treatment respectively \citep{Margulescu2017PersistentAF}. Although there has been extensive research on how to detect AF with high accuracy \citep{Asgari2015AutomaticDO,Larburu2011ComparativeSO,Martis2013AutomatedDO}, it remains a challenge to complete the discrimination of $N$, $AF_f$, and $AF_p$ types, and one step further, the localization of episodes of abnormal heart rhythms for $AF_p$ signals. To improve the practical value of automatic ECG diagnostic algorithm, we combine the three-class discrimination task with detailed localization task, closer to the realistic multi-task scenario. 

Typical methods of ECG signal diagnosis involve preprocessing, feature extraction, and classification based on machine learning algorithms, among which feature extraction often serves as the bottleneck of model improvement \citep{Jambukia2015ClassificationOE}. \cite{Zhao2005ECGFE} and \cite{Castro2000ECGFE} extract the coefficients of wavelet transform as features. \cite{Tadejko2007MathematicalMB} use morphological filters, as nonlinear signal transformations, to gather geometrical features. \cite{SlidingWindow2022} divide the raw sequences into RR interval sequences through sliding window and manually extract 12 features in time domain. However, such hand-operated feature engineering tends to be based on particular domain knowledge and specific data patterns, inevitably introducing undesirable prior bias. Further, many features require knowledge of signal theory or cardiology to understand, which can be an obstacle to practical use. This also poses limitation on extension of such models to similar scenarios dealing with sequential data.

Nowadays, the success of deep learning across multiple domains has inspired the utilization of various deep learning techniques in ECG. It has stronger representation capability to capture the dynamic changes in ECG signals and demonstrates greater AF detection performance \citep{Liu2018MultiplefeaturebranchCN}. Models based on CNN, Recurrent Neural Network (RNN) or a hybrid of the two have been proposed for AF detection. These networks enable the use of raw ECG signals as unstructured data and extract features automatically, thus reducing manual bias and often reporting better performance \citep{FAUST2018327, Hybrid2021, STFT2019}. Recent progress turns the spotlight onto temporal models and attention mechanism. \cite{Yu2019Lstm} show the Long and Short-Term Memory (LSTM) neural networks can accurately predict changes in time-series data with periodicity, while \cite{hu4098696detection} propose a deep learning model based on residual blocks and the Transformer \citep{vaswani2017attention} encoder to locate AF episodes, obtaining satisfactory performance. 

In terms of research involving the $AF_p$ type, there have been several methods performing well for both discrimination and localization tasks \citep{DetectionPAF2018, Multistep2022, TwoStep2022, WEN202219, petmezas2021automated}, but several limitations still follow. Firstly, many models \citep{Multistep2022, petmezas2021automated} start with traditional R-peak detection or RR interval segmentation. Nonetheless, this step itself is prone to noise and not accurate enough, giving rise to error accumulation. Secondly, the two-stage approach is commonly used \citep{TwoStep2022, bao2022paroxysmal}, with discrimination as the first stage followed by localization as the second stage. Essentially, these two tasks, focusing on different scales, should not be treated mutually independently. The two-stage framework fails on information integration and thus has space for performance lifting. Also, such strategy only performs localization based on instances predicted as $AF_p$ in the first stage. However, misclassification in the former stage is unavoidable, which will compromise the accuracy of episode localization in the second stage. Thirdly, some methods, involving esoteric model design and sophisticated post-processing, are not explicable and extensible for practical applications. For instance, \cite{WEN202219} propose $3$ neural networks and an auxiliary one for QRS detection to complete the two objectives separately. It's a quite intricate architecture with jargonistic merging procedure.




\begin{figure*}[!tpb]
\centerline{\includegraphics[width = 0.9 \textwidth]{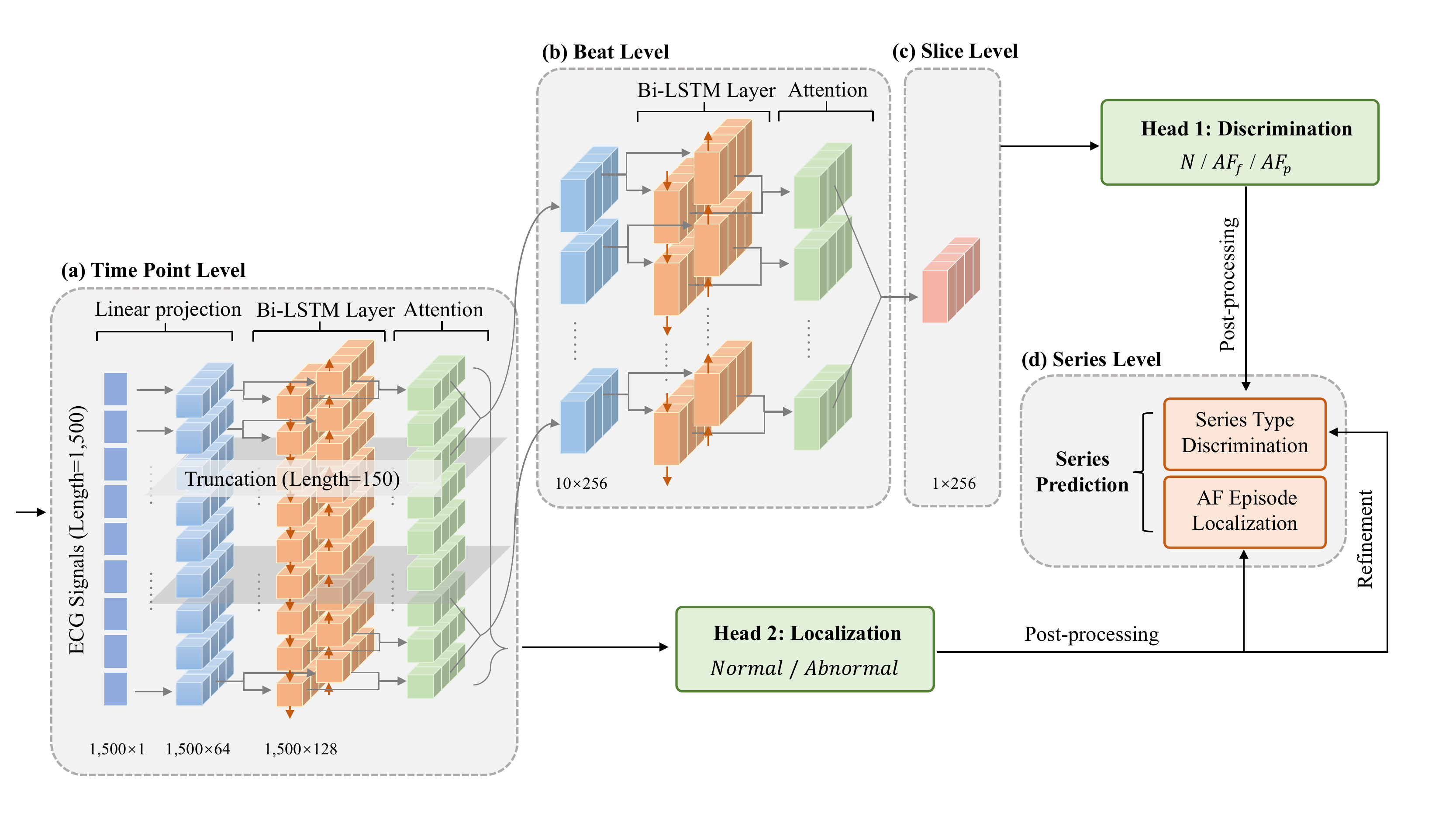}}
\caption{Framework of MMA-RNN. The MMA-RNN model contains three levels of feature extraction modules, with the architecture mainly consisting of Bi-LSTM and attention layers. Both Head $1$ and Head $2$ contain MLP layers, where the latter is a many-to-many structure.}\label{framework}
\end{figure*}

Accordingly, an integrated model which facilitates information interaction between these two objectives is urgently needed. Simplified and comprehensible preprocessing and post-processing steps are also expected. In this paper, we introduce the Multi-level Multi-task Attention-based Recurrent Neural Network (MMA-RNN) for discrimination and localization of AF based on ECG signals, which can adapt to variable-length signal inputs and enhance information sharing across different tasks. In order to leverage information at multiple levels, we extract features of time points, beats and slices. The low-level features extracted by point-based neural network, the medium-level features represented by beat-level truncation, and the high-level features captured by slice concatenation together generate a more integrated representation. Meanwhile, this hierarchical model is able to deal with unequal-length data without complex preprocessing, indicating greater data compatibility. We use CPSC 2021 \citep{Wang2021ParoxysmalAF} for training and testing the model. Experiments over this dataset show that our method can achieve promising performance in ECG diagnosis. It produces accurate results on both discrimination of three rhythm types and localization of abnormal episodes. Previously, \cite{mousavi2020han} propose a similar idea using Hierarchical Attention Network (HAN) to detect AF arrhythmia in an explainable way, but our model is completely different in pipeline design and implementation details, thus a novel framework focusing on $AF_p$ distinction problems with more pertinent design of inputs and multi-task training.

In conclusion, our main contributions are as follows:
\begin{itemize}
\item We provide a 
comprehensive solution for solving both discrimination of three types and localization of AF episodes, emphasizing the distinction between $AF_f$ and $AF_p$ patients and identifying the exact duration of AF episodes, which is of greater significance in clinical practice.

\item We propose an integrated end-to-end attention-based recurrent neural network for solving two tasks simultaneously with minimal manual bias, strong multi-level representation capabilities and less error accumulation, which can be implanted into wearable ECG devices. It also provides insights how to deal with interrelated tasks yet focusing on different 
scales.

\item Our experiments with CSPC 2021 dataset show that MMA-RNN has superior performance compared with existing methods on both two tasks. 

\end{itemize}

\section{Methods}

\subsection{Problem Formulation}
Given variant length ECG series $\{(s_{i1},\ldots,s_{it_i},y_i)\}_{i=1}^{N}$ sampled from $K$ patients, where $N$ denotes the number of ECG series, ${t_i}$ denotes the length of series $i$, $S = \{S_i|S_i = (s_{i1},\ldots,s_{it_{i}})\}$ denotes the series sample points and $y_i \in Y_s = \{N, AF_{f}, AF_{p}\}$ denotes the label of series $i$. Each patient may have multiple series. The discrimination task can be represented by a three-class prediction model $\phi: S \rightarrow Y_s$. For each series $S_i$, the accurate annotations of abnormal episodes, if any, are given. The localization task can be formulated as a binary classification problem. For each time point $s_{ik}, k = 1,...,t_i$ in the series, another model $\varphi(s_{ik})$ predicts whether it is normal or abnormal.

\subsection{Overview}

In this paper, we propose a novel architecture MMA-RNN for AF discrimination and localization. Figure \ref{pipeline} shows the pipeline of our study. Every raw series is cut into several fixed-length slices. The slices after normalization are input into our model without any other preprocessing. As for the model architecture, MMA-RNN is an end-to-end framework with a multi-level feature extraction module and two classification branches. In this way, it can model the hierarchical structure well and enable bidirectional information flow. The final results of two tasks for each series are obtained after simple post-processing.

\subsection{MMA-RNN}
\label{mma-rnn}
The architecture of MMA-RNN is shown in Figure \ref{framework}. Discrimination and localization focus on different granularity of information, with the former requiring high-level representation of slices and the latter using low-level representation of precise time points. Motivated by HAN \citep{HAN}, we propose a multi-level feature extraction module for the ECG series. HAN performs well in text classification tasks, successfully capturing the hierarchical structure of documents with multi-level attention mechanisms. In our network, we apply a three-level feature extraction of time point level, beat level, and slice level to obtain multi-grain information. Bottom-up feature aggregation is also achieved by this hierarchical architecture.

Furthermore, it is expected that the supervised information for each task can refine the feature representation of the corresponding level. Therefore, we design a two-head classifier. Head $1$ connects to the highest layer of feature extraction, and Head $2$ connects to the feature representation at time point level. 

Finally, it is necessary to train these two related tasks simultaneously rather than in stages. MMA-RNN is trained as an end-to-end framework to solve both tasks jointly, based on shared low-level features. In this way, we enable information to flow between two tasks.

\subsubsection{Multi-Level Feature Extraction Module}
In this section, we detail our three-level feature extraction for time point, beat, and slice level, respectively.

\textbf{Time point Level} The time point is the smallest unit in a slice and serves as the lowest level of our feature extraction module. A linear projection layer works for transforming the input of each time point into a high-dimensional vector with fixed size, to enhance the representation capability. Then, a Bi-LSTM layer is inserted to model the time points with sequential context from both directions. The final representations of time point level are obtained by concatenating the forward and backward hidden state vectors.

\textbf{Beat Level} The cyclicity of ECG signals is represented by heartbeats, which motivates the introduction of beat level. Features from the time point level are assembled in groups to generate beat-level inputs through the attention network. The attention weight $\alpha_{mt}$ of the $t$-th time point in the $m$-th beat is calculated as Equation \ref{att1} and Equation \ref{att2}.
\begin{equation}
q_{mt} = \textrm{tanh}(W_{B}h_{mt} + b_B),
\label{att1}
\end{equation}

\begin{equation}
\alpha_{mt} = \frac{\mathrm{exp}(q_{mt}^Tq_{B})}{\sum_{s}\mathrm{exp}(q_{ms}^Tq_{B})},
\label{att2}
\end{equation}

where $\{h_{mt}|m=1,\ldots,M,t=1,\ldots,T\}$ denotes the final representation vectors of the time point level. $W_B$ and $b_B$ are parameters and $q_B$ is a randomly initialized learnable context vector. $M$ is the total number of beats and $T$ is the number of time points in a single beat. Then, the $m$-th beat vector $c_m$ can be written as Equation \ref{att_sum1}.

\begin{equation}
c_m = \sum_{t}\alpha_{mt}h_{mt}.
\label{att_sum1}
\end{equation}

Given the beat vectors, we again employ a Bi-LSTM encoder to extract sequential information and attain the final representation $h_m$ of this level after concatenation.

\textbf{Slice Level} The slice level is the highest level in our feature extraction architecture, ultimately used for discrimination. The same attention method summarizes beat-level information and forms slice-level feature vectors. The attention weight $\alpha_m$ of the $m$-th beat is derived from Equation \ref{att3} and \ref{att4}.

\begin{equation}
q_{m} = \textrm{tanh}(W_{S}h_{m}+b_S),
\label{att3}
\end{equation}

\begin{equation}
\alpha_{m} = \frac{\text{exp}(q_m^Tq_{S})}{\sum_{k}\text{exp}(q_{k}^Tq_{S})},
\label{att4}
\end{equation}

The meanings of $W_S$, $b_S$ and $q_S$ here are similar to those of $W_B$, $b_B$ and $q_B$ in Equation \ref{att1} and \ref{att2}. Then we have the $n$-th slice representation vector $s$ by the weighted sum as Equation \ref{att_sum2}.
\begin{equation}
s = \sum_{m}\alpha_{m}h_{m}.
\label{att_sum2}
\end{equation}

\subsubsection{Multi-task Classifier}

MMA-RNN trains two tasks simultaneously. As for the discrimination task, Head $1$ makes predictions on the slice level. It takes the slice-level feature vectors as inputs and classifies each slice into three categories ($N$, $AF_f$, and $AF_p$) shown in Equation \ref{head1}. 
\begin{equation}
    \hat{y} = \textrm{softmax}(W_{d} s)
    \label{head1},
\end{equation}
where $W_{d}$ is the parameter of Multi-Layer Perceptron (MLP). As for the localization task, Head $2$ implements many-to-many binary classification on the time point level. It inputs time point representations $\{h_{mt}|m=1,\ldots,M,t=1,\ldots,T\}$, and then maps each time step vector to its prediction $\hat{z}_{mt}$ as Equation \ref{head2}. A dropout layer is added for regularization.
\begin{equation}
    \hat{z}_{mt} = \textrm{sigmoid}(\textrm{Dropout}(\textrm{MLP}(h_{mt})).
\label{head2}
\end{equation}

\subsubsection{Loss Design}
The loss function is defined as a weighted sum of mean cross-entropy losses from two tasks. For Head $2$, we calculate binary cross-entropy loss for each time point and then average all losses within each slice. The final loss function is as Equation \ref{loss_func}.
\begin{equation}
\mathrm{Loss} = w_d \times \mathrm{L_d} + w_l \times \mathrm{L_l}
\label{loss_func}
\end{equation}

where $\mathrm{L_d}$ and $\mathrm{L_l}$ denote the loss for Head $1$ and Head $2$. According to empirical results, We take $w_d = 1$ and $w_l = 40$ in subsequent experiments.
\begin{figure}[!tpb]
\centerline{\includegraphics[width=.5 \textwidth]{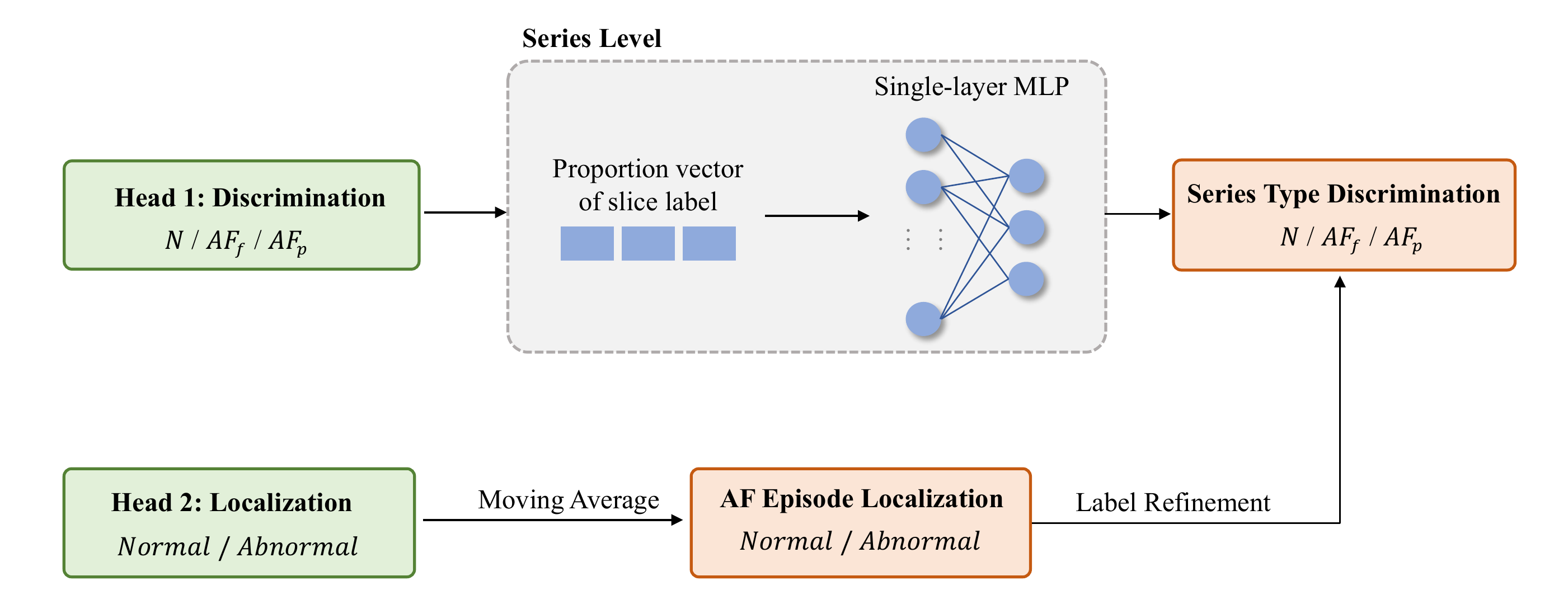}}
\caption{Steps of post-processing. The final results for both discrimination and localization are acquired from a few post-processing steps. A single-layer MLP is utilized to derive the type of each series from previously predicted labels of slices. A smoothing step calculating the moving average of binary predictions on time points refines the predictions of AF episodes. Finally, the results of Head $1$ and Head $2$ are integrated with simple rules.}\label{post-processing}
\end{figure}
\subsection{Post-processing}

\label{postproc}
So far, our model outputs class label predictions for each slice and binary label predictions for each time point in the slice. Here we illustrate our post-processing method to derive the final results, shown in Figure \ref{post-processing}.

We first convert the output of series $i$ from Head $1$ as a vector $\mathbf{p_i}=(p_{i1},p_{i2},p_{i3})^T$, where $p_{i1}$, $p_{i2}$ and $p_{i3}$ are the proportion of slices predicted as $N$, $AF_f$ and $AF_p$ in series $i$, respectively. Then we feed them into a single-layer MLP to conclude series-level labels. For the output of Head $2$, considering the continuity of both normal and abnormal episodes, we take a moving average with a fixed-length window to smooth the predicted results.

Lastly, we blend results from two tasks based on a straightforward rule. The series is simply categorized as $N$ (or $AF_f$) if Head $1$ labels it as $N$ (or $AF_f$). Otherwise, if there is a high ratio of time points in the series recognized from Head $2$ as normal (or abnormal), we will allocate it to $N$ (or $AF_f$), regardless of the judgement from Head $1$. The rest are classified to $AF_p$, and the localization results are dependent on Head $2$.

\begin{table*}
    \centering
    \small
    \caption{Data summary of the three datasets. Lengths are in the unit of seconds and rounded.}
    \label{dataset}
    \setlength{\tabcolsep}{2.7mm}{
        \begin{tabular}{l|cccccccc}
        \toprule
         Dataset & Size & $N$ & $AF_f$ & $AF_p$	& Avg. of length & Std. of length & Max. of length & Min. of length\\
         \midrule
Training &	$856$	& $50.58\%$ & $33.29\%$	& $16.12\%$	& $1,165$ & $2,022$ & $23,712$ & $8$\\
Validation	 & $285$	& $50.52\%$	& $33.33\%$ &	$16.14\%$ &	$1,311$ & $2,147$ & $24,667$ & $14$\\
Test & $284$ & $50.70\%$ & $33.45\%$ &	$15.85\%$ & $1,259$ & $1,743$ & $16,603$	& $15$\\
        \midrule
        \end{tabular}}
\label{datasummery}
\end{table*}

\section{Experiments and Results}

\subsection{Experimental setup}
\subsubsection{Datasets}
We apply the CPSC 2021 dataset for this study, which provides $1,425$ ECG records from $104$ patients with $3$ types ($N$, $AF_{f}$, $AF_{p}$). All records, sampled at $200\textrm{Hz}$ with significantly varying lengths, are collected from the lead I and lead II of 12-lead Holter or 3-lead wearable ECG devices. For the $AF_{p}$ series, each AF episode is annotated. Further, to avoid ambiguity, every AF and non-AF episode must contain no less than $5$ heart beats. In our experiments, the dataset is randomly divided into $3$ splits, in a stratified fashion, with $856$, $285$, and $284$ series for training, validation and test. Statistics of these datasets are summarized in Table \ref{datasummery}. We only use lead II in the following experiments. 


\subsubsection{Metrics}
\begin{figure}[!tpb]
\centerline{\includegraphics[width=0.25 \textwidth]{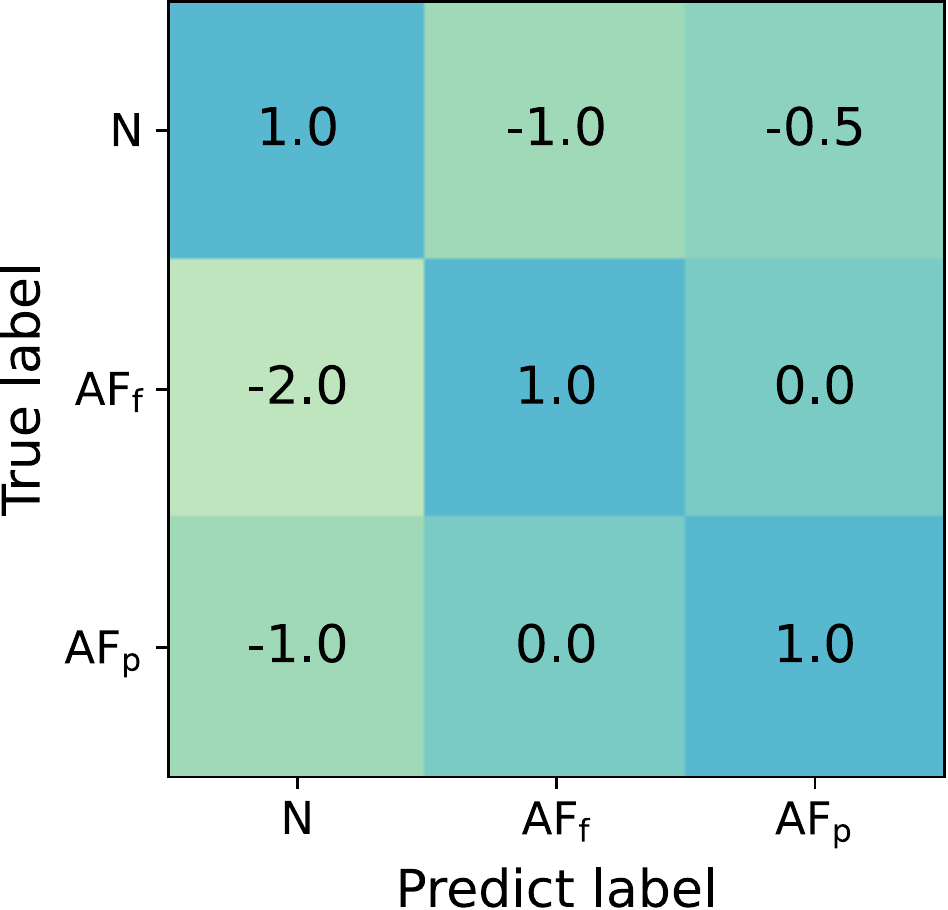}}
\caption{Scoring matrix of $U_r$ score. For example, if the prediction is correct, a score of $1.0$ is obtained. If a Normal person is classified as $AF_f$, a score of $1.0$ is deducted. If an $AF_f$ patient is identified as Normal, which is risky in clinical practice, a score of $2.0$ is deducted.}
\label{matrix}
\end{figure}

CPSC 2021 official evaluation protocol is adopted in our experiments. $U_{r}$ score assesses performance on the three-class discrimination task, which is calculated based on the scoring matrix presented in Figure \ref{matrix}, including weighted rewards and penalties. $U_{e}$ score evaluates the accuracy of episode detection. It is rewarded with $+1$ if the detected onset (or endpoint) is within $\pm1$ beat of the annotated position, and with $+0.5$ if it is within $\pm2$ beats. The final $U$ score is the sum of $U_{r}$ score and weighted $U_{e}$ score, averaged across all samples, defined as Equation \ref{u_score}.
\begin{equation}
U = \frac{1}{N}\sum\limits_{i=1}^{N}(U_{r_i}+\frac{M_{a_{i}}}{\max(M_{r_i},M_{a_i})}\times U_{e_i})
\label{u_score}
\end{equation}

Here $N$ is the number of records in the test set. $M_{a}$ and $M_{r}$ are the number of AF episodes from annotated answers and predicted results for each record, respectively. Such weight design is intended to penalize overestimating the number of AF episodes.

\subsubsection{Baselines}
Four baseline models (UNIWA, CUICM, Dabin and WHS) from the CPSC 2021's leaderboard are reproduced. They all employ CNN to extract morphological features within a fixed-length time window, followed by RNN to better use sequential information or attention mechanism to allocate focus.

\subsubsection{Implementation details}
For MMA-RNN, we set the length of beats and slices to $150$ and $1,500$. The former is based on the fact that normal resting heart rate is between 60 and 100 per minute in adults \citep{o2018beats}, and the latter is taken roughly $10$ beats. The number of units for the linear projection layer is chosen as $64$. As for Bi-LSTMs, we apply one layer where the dimensionality of latent spaces is set to be $128$. As for the classification heads, each slice is labeled with $3$ possible values according to the series type, while each time point is labeled in a binary manner. Dropout rate in all MLPs is set to be $0.5$ and Xavier initialization \citep{he2015delving} is utilized. Regarding post-processing, we use a simple $100$-unit single-layer MLP on the output of Head $1$ to obtain predicted series labels. Moving Average modification with a window size of $1,200$ is adopted on the output of Head $2$. The window size is determined in keeping with the instructions of the database.

For the baseline models, we train each of them referring to the learning curve on validation set. Each model is trained until convergence.

\begin{figure*}[!htb]
    \centering
    \subfigure[Model performance on the validation split]{\includegraphics[width=0.45 \textwidth]{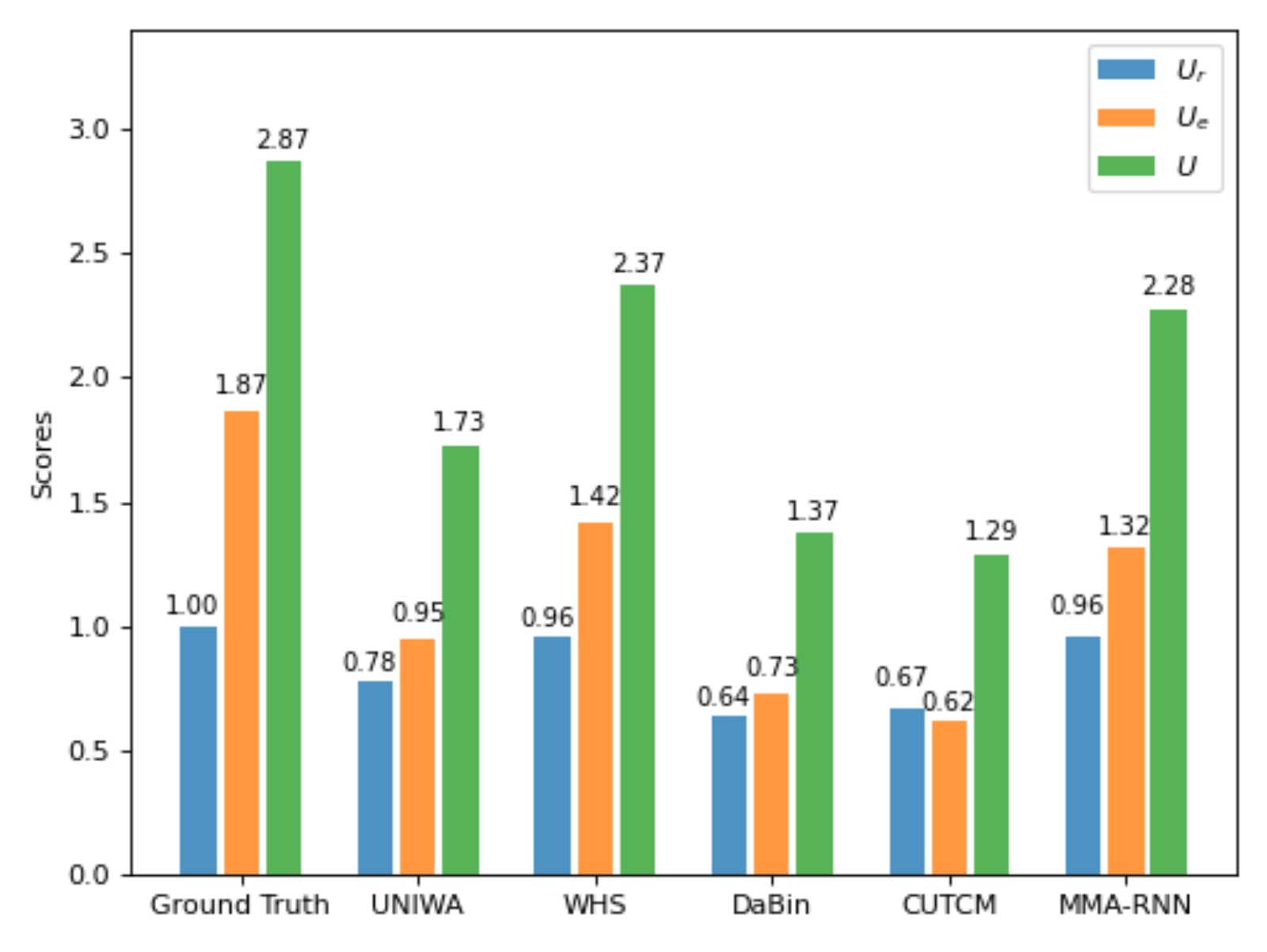}
        \label{comparison1}}
    \quad
    \subfigure[Model performance on the test split]{\includegraphics[width=0.45 \textwidth]{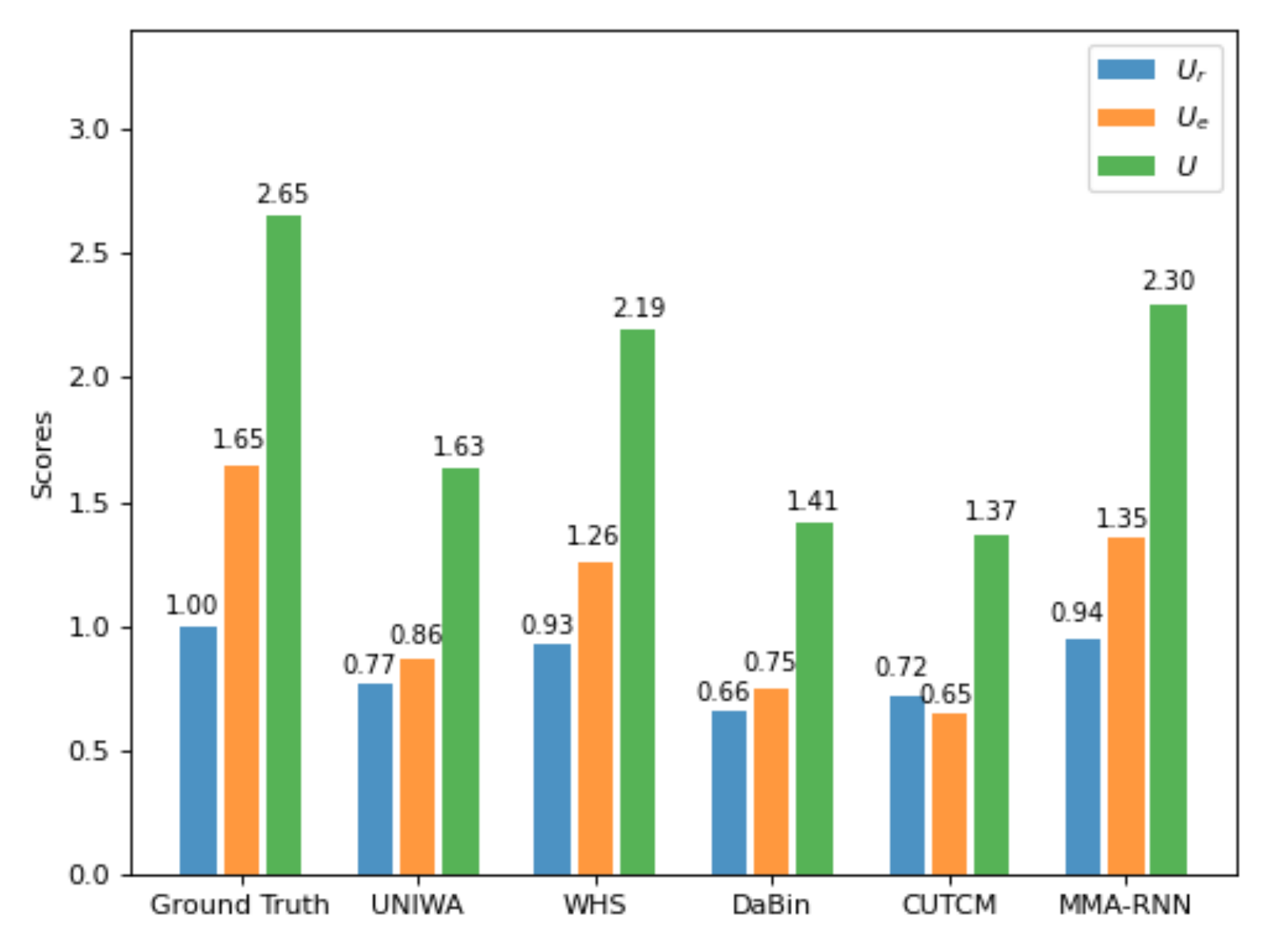}
        \label{comparison2}}
    \caption{Model performance on CPSC 2021 dataset. (a) shows the results on the validation split compared with four baselines and ground truth. (b) shows the results on the test split compared with four baselines and ground truth. MMA-RNN exceeds all the baselines and demonstrates stable performance.}
    \label{comparison}
\end{figure*}

\begin{figure*}[!tpb]
\centerline{\includegraphics[width=1.0 \textwidth]{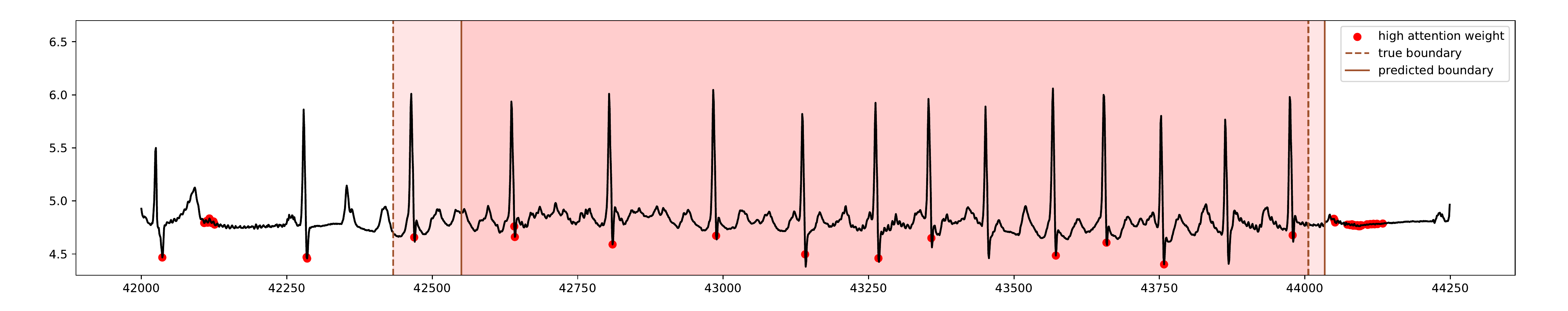}}
\caption{Case Visualization. In this figure, an episode from one $AF_p$ signal is presented. The dashed lines represent the true onsets and end points of $AF_p$ episodes. The solid lines represent the predicted boundaries, with the overlapping area shaded in deeper red. It can be observed that our localization predictions are close to the true boundaries. Time points with larger attention weights are marked by red dots.}
\label{attplot}
\end{figure*}

\subsection{Performance evaluation}

\subsubsection{Results}
Figure \ref{comparison} reports results on validation and test split. On the test set, our MMA-RNN shows the best performance in terms of $U$ score, achieving $2.2975$ and exceeding all other baselines by a large margin. A $U_{r}$ score of $0.9437$ out of $1$ suggests the great ability for discrimination, and a $U_{e}$ score of $1.3538$ out of $1.6479$ indicates our method can locate the onsets and end points precisely. 

For each given metric, our model markedly outperforms baseline models UNIWA, CUICM and Dabin, on both validation and test split. It demonstrates that our model can perform stably well. Our model exceeds WHS over $0.1$ points on the test set. WHS is the only one applying attention mechanism among all the baselines. This implies the necessity of attention design.

\subsubsection{Visualizations}

We visualize the results here to dig deeper into the potential of our method. A case of localization is given in Figure \ref{attplot}. It can be observed that the predicted boundary is close to the true boundary, with the margin of error less than one beat. Figure \ref{attplot} also marks the time points with high attention score. These points are distributed across beats, indicating although the slices are fed as the input, the attention mechanism still manages to identify each single beat. Consequently, each single beat can provide valuable information to our model. Some points scatter around the predicted end point, suggesting the model indeed learns that more attention should be allocated to the boundary of AF episodes.


\subsection{Model Analysis}
In this section, we take a closer inquiry on our model to further demonstrate the effectiveness of module design. The following results are discussed on the test split.
\subsubsection{Architecture analysis}
We would like to find out whether the information interaction between various levels is indispensable, and in that case, how to design the interaction procedure. Specifically, we design the following three modeling schemes. 

\begin{table*}
    \centering
    \small
    \caption{Results of ablation experiments for model architecture on the test set. Experiments with different levels and different ways of information interaction are presented, evaluated by $U_r$, $U_e$	and $U$ scores.}
    \label{3step}
   \setlength{\tabcolsep}{7mm}{
        \begin{tabular}{ccc|ccc}
        \toprule
         High-Level & Low-Level 	& Interaction & 	$U_r$ & $U_e$	& $U$\\
         \midrule
$-$	& \checkmark	& $-$&	$0.8204$ &	$1.1864$ & $2.0069$ \\
\checkmark &\checkmark &$-$& $0.9208$ & $1.3059$ & $2.2267$\\
\checkmark & \checkmark & Pre-train $1$; Fine-tune $2$ & $0.9263$ & $\mathbf{1.3542}$ & $2.2805$\\
\checkmark & \checkmark &	Pre-train $2$; Fine-tune $1$	& $0.9085$	& $1.2927$ & $2.2011$ \\
\checkmark &	\checkmark &	MMA-RNN & $\mathbf{0.9437}$ & $1.3538$ & $\mathbf{2.2975}$ \\
        \bottomrule
        \end{tabular}
    }
\end{table*}
\begin{itemize}
  \item {\textbf{Only low-level features:} We remove Head $1$ and make predictions entirely based on Head $2$.}
  \item {\textbf{Both low-level and high-level features, no information interaction:} We split our multi-task architecture into two mutually independent networks.}
  \item {\textbf{Both low-level and high-level features, pre-training and fine-tuning as information interaction:} We remove Head $2$, train the network and get the output of Head $1$. Then we discard Head $1$, and bring back Head $2$ to fine-tune the pre-trained network. We also switch the role of Head $1$ and Head $2$. Consequently, the information flow is uni-directional, thus the information interaction exists, but in a deficient way.}
\end{itemize}
Results are presented in Table \ref{3step}. It indicates that when information from multiple levels interacts in an elaborate way, the performance of both tasks tends to lift. The success that MMA-RNN achieves on synergy of information is also explicitly shown.

Another ablation experiment (Table \ref{SliceFeature}) demonstrates the importance of introducing individuality. Instead of only feeding every time point representation into Head $2$, we concatenate it with the slice-level feature representation and then perform many-to-many classification on these longer representation vectors. However, as shown in Table \ref{SliceFeature}, the performance even drops when more information sharing is brought in. We conclude that the information exchange should be conducted moderately, neither too little nor too much.

\begin{table}
    \centering
    \small
    \caption{Ablation experiment results on with / without slice-level features on Head $2$, evaluated by $U_r$, $U_e$, and $U$ score on the test split.}
    \label{SliceFeature}
    \setlength{\tabcolsep}{4.55mm}{
        \begin{tabular}{c|ccc}
        \toprule
        Slice Features & $U_r$ & $U_e$ & $U$\\
        \midrule
w/ & $\mathbf{0.9437}$ &	$\mathbf{1.3538}$ & $\mathbf{2.2975}$ \\
w/o & $0.9208$ & $1.2591$ & $2.1799$ \\

        \bottomrule
        \end{tabular}
    }
\end{table}

\begin{table}
    \centering
    \small
    \caption{Ablation experiment results on dimensionality ($d$) in the linear projection layer, evaluated by $U_r$, $U_e$, and $U$ score on the test split.}
    \label{LinearProjection}
    \setlength{\tabcolsep}{4.55mm}{
        \begin{tabular}{c|ccc}
        \toprule
        \ $d$ & $U_r$ & $U_e$ & $U$\\
        \midrule
$0$ &  $0.9173$ &	$\mathbf{1.3771}$ &	$2.2944$\\
$8$ &	$0.9349$ &	$1.3433$ &	$2.2782$ \\
$64$ &	$\mathbf{0.9437}$ & $1.3538$ & $\mathbf{2.2975}$\\
$256$ &	$0.9243$ &	$1.3323$ &	$2.2566$\\

        \bottomrule
        \end{tabular}}
\end{table}

\begin{table}
    \centering
    \small
    \caption{Ablation experiment results on slice length, evaluated by $U_r$, $U_e$, and $U$ score on the test split.}
    \label{SliceLength}
    \setlength{\tabcolsep}{4.55mm}{
        \begin{tabular}{c|ccc}
        \toprule
        Slice Length &  $U_r$ & $U_e$ & $U$\\
        \midrule
$150$ & $0.8838$ & $1.3024$ & $2.1862$\\
$1,500$ & $\mathbf{0.9437}$ & $\mathbf{1.3538}$ & $\mathbf{2.2975}$\\
$6,000$ & $0.9155$ & $1.2777$ & $2.1932$\\
        \bottomrule
        \end{tabular}
    }
\end{table}

\subsubsection{Parameter sensitivity}
Table \ref{LinearProjection} and Table \ref{SliceLength} present our further study on ablation of important parameters.
\begin{enumerate}
    \item {\textbf{Linear Projection Layer Dimensionality}} This layer is added to infuse each time point with richer information. Table \ref{LinearProjection} shows that $U_r$ rises while $U_e$ drops after applying this layer. The reason behind is when more information is available, it demands a deeper network to learn meaningful representations. Therefore, the shallow Head $2$ tends to perform a bit poorly while the high-level Head $1$ is able to produce better results. Overall, the introduction of this layer lifts the final performance. 
    \item {\textbf{Slice Length}} We introduce the slice level to guarantee that the information is aggregated in a subtle way. When this level is either removed or the slice is too lengthy, the performance also suffers.
\end{enumerate}

\section{Conclusion}

This paper focuses on studying the challenging AF detection problems for classifying ECG signals into $\{N, AF_{f},$ $ AF_{p}\}$ and further localizing the onsets and end points of AF episodes. To this end, we propose the MMA-RNN model, an end-to-end structure tackling the two tasks simultaneously. Our method, orchestrating information exchange in an elegant way, circumvents complex preprocessing and post-processing, and enjoys less error accumulation and greater data compatibility. MMA-RNN composes of two novel designs: a hierarchical feature extraction module to gather multi-level information from raw data and a multi-task strategy to promote information synergy between different tasks. Effectiveness of our network architecture is validated by extensive experiments on CPSC 2021 dataset. 

As for clinical applicability, real-time and dynamic monitoring of AF helps routine care and diagnosis, allowing for timely initiation of therapies and prevention of complications. Our economical end-to-end structure can be well inserted into wearable devices, which is of great value for AF early identification. As for extension, the proposed model can be transferred into other signal processing scenarios with similar data structures, such as tasks on electroencephalography (EEG). It also provides insights on how to deal with double tasks for discrimination and localization, such as motif identification of gene expression, in a more effective way.

\section*{Declaration of interests}
The authors declare that the research was conducted in the absence of any commercial or financial relationships that could be construed as a potential conflict of interest.

\section*{Data and codes availability}
The data and source code are available online at \url{https://github.com/JYS-99/MMA-RNN}.

\section*{Acknowledgement}
This work is partially supported by NSFC grant 61721003. 










\printcredits


\bibliographystyle{cas-model2-names}


\bio{}
\endbio

\endbio

\end{document}